# Perimeter Control with Heterogeneous Metering Rates for Cordon Signals: A Physics-Regularized Multi-Agent Reinforcement Learning Approach

Jiajie Yu[1,2,*], Pierre-Antoine Laharotte[2], Yu Han[3], Wei Ma[1], Ludovic Leclercq[2]

1. Civil and Environmental Engineering, The Hong Kong Polytechnic University, Hung Hom, Kowloon, Hong Kong Special Administrative Region, China

2. Univ Gustave Eiffel, ENTPE, LICIT-ECO7, F-69675 Lyon, France

3. School of Transportation, Southeast University, 21189 Nanjing, China

**Abstract:** Perimeter Control (PC) strategies have been proposed to address urban road network control in oversaturated situations by regulating the transfer flow of the Protected Network (PN) based on the Macroscopic Fundamental Diagram (MFD). The uniform metering rate for cordon signals in most existing studies overlooks the variance of local traffic states at the intersection level, which may cause severe local traffic congestion and degradation of the network stability. PC strategies with heterogeneous metering rates for cordon signals allow precise control for the perimeter but the complexity of the problem increases exponentially with the scale of the PN. This paper leverages a Multi-Agent Reinforcement Learning (MARL)-based traffic signal control framework to decompose this PC problem, which considers heterogeneous metering rates for cordon signals, into multi-agent cooperation tasks. Each agent controls an individual signal located in the cordon, decreasing the dimension of action space for the controller compared to centralized methods. A physics regularization approach for the MARL framework is proposed to ensure the distributed cordon signal controllers are aware of the global network state by encoding MFD-based knowledge into the action-value functions of the local agents. The proposed PC strategy is operated as a two-stage system, with a feedback PC strategy detecting the overall traffic state within the PN and then distributing local instructions to cordon signals controllers in the MARL framework via the physics regularization. Through numerical tests with different demand patterns in a microscopic traffic environment, the proposed PC strategy shows promising robustness and transferability. It outperforms state-of-the-art feedback PC strategies in increasing network throughput, decreasing distributed delay for gate links, and reducing carbon emissions.

## 1. Introduction

Traffic signal control is vital in urban traffic network control, directly affecting traffic efficiency, network stability, and vehicle emissions (Han et al., 2016; Yu et al., 2023). Various advanced adaptive traffic signal control strategies have been explored to alleviate congestion, considering scalable frameworks, multi-modal networks, and autonomous vehicle environments (Chen et al., 2022b; Wang et al., 2021; Yu et al., 2023). However, for dense road networks, such as urban grid networks, local traffic signal control strategies may lead to gridlocks when the network is oversaturated (Laval and Zhou, 2022). Hence, Perimeter Control (PC) was introduced to protect a specific urban region inside a cordon, also called the Protected Network (PN) (Geroliminis et al., 2012; Haddad and Geroliminis, 2012). Normally, PC strategies regulate the transfer flow of PN through the metering policy of cordon signals or gates, aiming to maximize the outflow of PN by keeping the vehicle accumulation inside PN in the critical range of the Macroscopic Fundamental Diagram (MFD) (Geroliminis et al., 2012). This maintains the network throughput at its maximum level and prevents gridlocks from happening (Haddad and Mirkin, 2020).

PC strategies with uniform metering rates for gates located in the same perimeter have been extensively explored with Model Predictive Control (MPC), feedback Proportional–integral (PI) control, data-driven approaches, and Reinforcement Learning (RL) techniques. Specifically, the MPC models the traffic dynamics with MFD and obtains the optimal PC policy (i.e., metering rate of cordon signals) over a prediction horizon (Geroliminis et al., 2012; Haddad, 2017; Haddad et al., 2013; Ramezani et al., 2015). Without full MFD estimations, PI control strategies generate the real-time transfer flow limit for cordon gates according to the current, previous, and critical total time spent inside the PN at every control step (Keyvan-Ekbatani et al., 2012; Kouvelas et al., 2015). Alternative data-driven PC strategies were also proposed to achieve adaptive PC. For example, an adaptive PC strategy for a multi-region network depending on real-time accumulation and PC input is developed by Haddad and Mirkin (2017). Lei et al. (2020) and Hou and Lei (2020) considered route guidance in an adaptive PC strategy for a multi-region network, using the real-time traffic data (e.g., input and output data) of the network to replace the traffic dynamics model. As the Reinforcement Learning (RL) technique enables traffic control, a set of studies applied two-region and multi-region PC based on RL frameworks, showing comparable performance to model predictive control and high transferability. In Su et al. (2023), Zhou and Gayah (2021), and Zhou and Gayah (2023), each perimeter of PN is regarded as an RL agent learning the metering policy, and all gates in the same perimeter act with a uniform metering rate. In Chen et al. (2022a), the agent is designed to learn the macroscopic urban traffic dynamics for multi-region networks, and then a uniform metering rate is calculated for every perimeter based on the learned macroscopic traffic models. Compared to model-based PC strategies, RL techniques have the potential to enhance PC strategies with model-free and responsive properties.

The aforementioned studies have validated that PC strategies can effectively prevent PN from being oversaturated. However, the uniform metering rate might be unfair for the gates with higher demand, which may cause severe traffic delays at the related cordon intersections (Keyvan-Ekbatani et al., 2021).

Some studies conduct PC with heterogeneous metering rates for cordon signals as centralized controllers based on PI control and RL framework. For example, the PC strategies in Keyvan-Ekbatani et al. (2021) and Tsitsokas et al. (2023) balance the cordon queue by allowing larger inflow volumes for the gates with higher demand under the premise of the total inflow limit from the classical PI strategy. Ni and Cassidy (2019) applied the RL framework to PC with spatially varying metering rates. The perimeter is regarded as a centralized RL agent that chooses different metering rates for all cordon signals according to the local accumulations and the metering rate resulting from the uniform PC policy. Findings highlight that the vehicle hours traveled is reduced in this strategy compared to the uniform PC strategies. These studies have revealed the significance of considering both the network's global state and local traffic conditions along the cordon during the process of PC. However, the complexity or agent's action space in these centralized frameworks increases exponentially with the network size, resulting in diminished scalability and reduced transferability of the strategies.

To our best knowledge, decentralized or distributed formulations of PC with heterogeneous metering rates for cordon signals were only explored via Nash bargaining optimization in the literature. Elouni et al. (2021) implement a PC strategy through game-theoretic concepts based on Nash bargaining optimization. The decentralized controller activates the phase, including the PC phase (i.e., closing the cordon gate), that minimizes queue length for each cordon intersection at every step. It outperforms the centralized PC on traffic delay and carbon emission, highlighting the benefits of scalability and transferability to PC strategies without compromising performance. However, this decentralized method provides optimal solutions for the current traffic state at each time step without ensuring the long-term control effect.

Regarding PC strategies considering heterogeneous metering rates, cordon signals act separately and cooperate in a shared environment. This strategy is suitable to be implemented with a Multi-Agent Reinforcement Learning (MARL) framework, in which each agent controls an intersection in the cordon and contributes to the common purpose of protecting the perimeter. The long-term performance of the network system can also be considered in the MARL framework. Although the MARL framework can bring the distributed property to PC, the local signal control-based strategy without considering the global network state may still lack stability in oversaturated situations (Fu et al., 2017). To address this issue, PC strategies can provide physical knowledge based on MFD to instruct cordon controllers. Given this, the cooperation of the network-level PC with distributed signal control at the local intersection level can be achieved by incorporating physics into the learning-based method.

Physics-informed and physics-regularized approaches have been proposed in the literature to leverage physical knowledge to improve the performance of learning-based methods. For example, Mo et al. (2021) encoded physics-based car-following models into neural network architectures, forming a physics-informed deep learning framework for car-following modeling. The method shows high estimation accuracy, and the training process is also data-efficient. Shi et al. (2023) and Han et al. (2022) proposed physics-informed RL methods that guide the RL agents' exploration with model-based knowledge (e.g., traffic flow models) to enhance training efficiency. Yuan et al. (2021) proposed a physics regularization method for the Gaussian process in traffic flow modeling. The method encodes macroscopic traffic flow model-based knowledge into the stochastic process, showing better estimation precision and robustness to noisy training datasets. Yuan et al. (2022) further introduced a gradual physics regularized learning method to improve the efficiency of previous work. Zheng et al. (2024) developed a doubly physics-regularized denoising diffusion model for traffic data recovery from irregular noise, able to provide an unbiased estimation of the true value under suitable assumptions. These studies bring insights to this paper that encode the MFD-based knowledge to the MARL-based cordon signal control framework.

To summarize, PC with heterogeneous metering rates for cordon signals is complicated for centralized controllers since each signal acts differently leading to an extremely large action space. The decentralized or distributed formulations of the problem have not been sufficiently explored, and the interaction between the strategies from two layers (the PC at the network level and the local cordon signal control at the intersection level) needs to be well-designed. Therefore, this study leverages the MARL-based traffic signal control framework to decompose the PC problem into multi-agent cooperation tasks to address different local traffic states at cordon intersections. A novel physics regularization method is introduced for the MARL framework to interact distributed traffic signal control with feedback PC. The MFD-based knowledge regularizes MARL agents to consider the PC policy while conducting individual local signal control. This paper's primary contributions are summarized below:

• This paper formulates the PC problem considering heterogeneous metering rates for cordon signals into a MARL framework. The distributed property of the MARL framework makes the precise control of cordon signals easier than centralized formulations due to the reduction of the problem's complexity.

• This paper proposes a physics regularization approach for the MARL framework. This approach encodes the physical knowledge from MFD into the action-value function of MARL agents, effectively utilizing the traffic flow models to instruct the behaviors of MARL agents.

• The transferability of the proposed method is demonstrated by inheriting each of the trained models to multiple agents (i.e., cordon signal controller) during tests without retraining. Both the global throughput of the PN and the distributed delay of each cordon gate are verified as promising in the proposed method via tests.

The remainder of the paper is organized as follows. Section 2 introduces the classical PI control strategy which inspires the formulation of PC feedback in the upper layer (global perspective) of our method and states the problem addressed by this paper.

Section 3 first describes the physics-regularized MARL framework, followed by the local controller design based on MARL as the lower layer (local perspective) and the physics regularization for agents' action-value function. The microscopic simulation tests with different traffic demand patterns are displayed and discussed in Section 4. The conclusions and perspectives are summarized in Section 5.

## 2. Problem formulation

In the PN (e.g., the city center) depicted by the shadow area in Fig. 1, the PC strategy is applied to the signals located at the cordon of the PN (highlighted with blue dots in Fig. 1) to regulate the accumulations inside the PN by metering the transfer traffic volume along the cordon. Classical PI control calculates the permitted total inflow volume of the PN for each control step according to the current Total Travel Time (TTT), last-step control information, and critical TTT of MFD (Keyvan-Ekbatani et al., 2012). The total inflow evolution of PN can be formulated as Eq. (1).

$$q_g(k) = q_g(k-1) - K_P\left[TTT(k) - TTT(k-1)\right] + K_I\left[\hat{TTT} - TTT(k)\right] \tag{1}$$

where $q_g(k)$ is the total flow entering the network through the gated links, and $TTT(k)$ is the TTT inside the PN at time step $k$. $K_P$ and $K_I$ are the proportional and integral gains, which need to be calibrated with the control network. $\hat{TTT}$ is the critical TTT of MFD within the PN. The calculation of TTT and detailed formulation of PI control are available in (Keyvan-Ekbatani et al., 2012) and (Keyvan-Ekbatani et al., 2021).

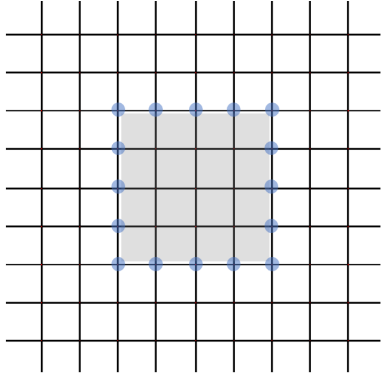

Fig. 1. Two-region grid network

The classical PI control involves a uniform distribution of the metering rate among all cordon signals, disregarding the heterogeneous traffic demand at each gate. To balance the cordon queue length, PI control considering the cordon queue (denoted as "PI-cordon queue") formulated by Keyvan-Ekbatani et al. (2021) distributes the inflow volume to each gate by solving an integer quadratic programming problem. More detailed formulations of PI-cordon queue are concluded in Appendix A.1. This approach seeks to allocate a larger inflow to the gate with a larger relative queue length. However, a large inflow demand at a gate might easily cause a local overload of the links in the surroundings, then achieving a local gridlock. The traffic distribution inside the PN would be heterogeneous, leading to network throughput drops that degrade the network stability. Although the cordon queue length can be balanced in the short term, the long-term performance of the network might be affected by the throughput drop. Theoretical analysis to illustrate this hypothesis can be found in Appendix A.2. Therefore, different methods to refine the inflow distribution for the PC need to be explored to consider the long-term system effectiveness and allocate the inflow to each gate wisely. Precisely, RL techniques are ideal for incorporating the environment's long-term feedback to guide agents' behavior incrementally (Mnih et al., 2013; Mnih et al., 2015).

In this paper, the proposed method is supposed to have the following features: each cordon signal acts to maintain the traffic system efficiency at the intersection level and in its surroundings, and when a PC is needed (i.e., the PN is oversaturated), cordon signals will mitigate between the original local action and the global recommendation from the network level (i.e., the PC strategy). To ensure the scalability and transferability of the method, the MARL framework is used to build the cordon signal controllers. Since the evolution of global traffic states (e.g., accumulations and TTT inside the PN) due to the cordon signals' action has been explicitly formulated by MFD and feedback PC strategies, it does not need exploration through the data-driven approach such as MARL. By removing the global traffic control from the learning objective of MARL agents, their state space can also be reduced significantly which may benefit their scalability and training efficiency (Tesauro et al., 2006; Zhang et al., 2023). Therefore, we proposed a physics-regularized MARL framework to separate the agent training process and the traffic model encoding process. The agents in the MARL framework consider only local traffic state at the local intersection level. The global network state is integrated with the action-value function of each agent to regularize agents' behavior only in the testing process. The detailed descriptions and formulations of the framework are introduced in Section 3.

## 3. Perimeter control with heterogeneous metering rates for cordon signal

When implementing perimeter control, it is essential to consider both the cordon traffic state and the network's overall state. In the proposed framework, the PC feedback is the difference between current TTT and critical TTT inside the PN. This formulation of PC feedback is inspired by PI control, as PI control does not require a full-estimated MFD and shows better robustness in unestimated traffic conditions compared to model predictive control (Han et al., 2020; Keyvan-Ekbatani et al., 2015a). This combination of local cordon signal control and PC feedback is executed by the physics-regularized MARL framework.

### 3.1. Overview framework of physics-regularized MARL

The structure of the physics-regularized MARL framework is illustrated in Fig. 2. Each agent in the MARL framework represents a controller for an individual cordon signal. The training and testing processes of these agents are separated. The agents are offline trained solely based on traffic states at the local intersection level and its surroundings. In the testing process, the proposed framework utilizes PC policy to provide global feedback (i.e., MFD-based knowledge) to regularize the local controllers' actions for cordon signals.

When the TTT inside the PN is larger than the critical TTT during the testing process, the cordon signals are supposed to activate the PC phase, which closes the inflow gates of the PN to decrease the TTT inside. From the perspective of the cordon signal's local control, if the action value of the PC phase is higher than the one associated with alternative actions, then the PC phase will be selected directly. However, if the values of the alternative actions from the local control agent remain higher than the action suggested by the PC, then a tradeoff must be made between local control and the PC. In the forthcoming section, we outline the details of the local control agents' design and the physics regularization for the action-value function of MARL agents.

The proposed local cordon signal control is a distributed system that consists of two types of cordon signals denoted as "trained CSs" and "inheriting CSs". The trained CSs are effectively trained with local MARL agents, while the inheriting CSs directly inherit knowledge from the trained ANNs during testing. To cover different road configurations, a set of trained CSs is selected, and then their knowledge is transferred to the inheriting CSs based on the shared road configuration. In Fig. 2, the cordon signals with the same marked color share the same trained ANN. When the cordon of PN changes, the trained CSs can still be transferred to the new cordon signals without retraining if the intersection configuration (e.g., number of incoming legs, number of phases) is the same. Only the critical TTT of the PN needs to be recalibrated in the testing process. This ensures the transferability of the trained agents.

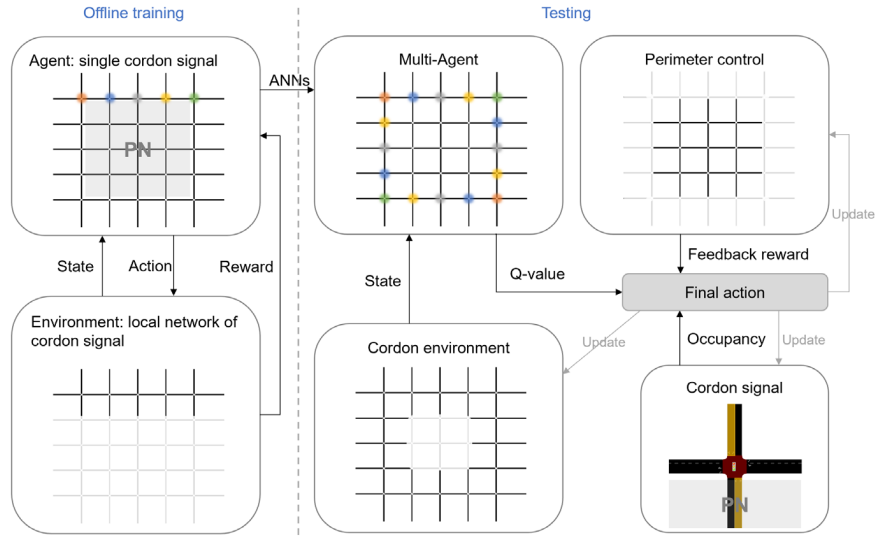

Fig. 2. Structure of physics-regularized MARL framework.

### 3.2. Local control for cordon signals

RL techniques have demonstrated high efficiency in traffic signal control with responsiveness, scalability, and transferability (Chu et al., 2019; Han et al., 2023; Yu et al., 2023). The proposed strategy uses a MARL framework to build the local control system where each cordon signal is regarded as an independent agent. All agents operate in the same shared road network, and the Double Deep Q-Network (DDQN) is applied to the learning process (Van Hasselt et al., 2016). During the training process of local cordon signal control, only the local traffic state at the intersection level is observed, and the network global state (e.g., accumulation or TTT within the PN) is not taken into account. The value-based RL algorithm is selected as it evaluates each possible action for its value at every action step, with the action value representing the long-term reward of the related action. This allows the local control reward coupled with the PC feedback to regularize the action-value function during testing.

**1) Action-value function approximation: DDQN**

DDQN is proposed to solve the over-optimization issue in Deep Q-Network (DQN) caused by the Q-network overestimating the action values (Van Hasselt et al., 2016). In DDQN, there are two Q-networks (an online network and a target network). $\theta$ and $\theta'$ are the weights of the two Q-networks represented by two Artificial Neural Networks (ANNs). The action selection is according to the Q-value estimated by the online network, while the target network evaluates the Q-value of the selected action in the error update. The weights of the online network $\theta$ are updated after each learning, which is the same as DQN. The target network stays unchanged and only updates its weights $\theta'$ with a predefined frequency. The target in DDQN is defined as Eq. (2).

$$Y_t \equiv R_{t+1} + \gamma Q(S_{t+1}, \arg\max_a Q(S_{t+1}, a; \theta_t); \theta'_t) \tag{2}$$

where $R_{t+1}$ is the immediate return at the current time step $t$. $\gamma$ is the discount factor of future reward. $S_{t+1}$ is the state at time step $t+1$. $Q(s, a; \theta_t)$ is the Q-value function with state $s$, action $a$, and weights $\theta_t$. Then, the weights of online networks will be updated as Eq. (3).

$$\theta_{t+1} = \theta_t + \alpha(Y_t - Q(S_t, A_t; \theta_t))\nabla_{\theta_t} Q(S_t, A_t; \theta_t) \tag{3}$$

where $\alpha$ is a scalar step size. The weights of the target network are replaced by $\theta_t$ every $N^-$ learning iteration, where $N^-$ is the target network replacement frequency. More details about DQN and DDQN are available in (Mnih et al., 2013; Mnih et al., 2015; Van Hasselt et al., 2016).

**2) Agent's design**

Each agent represents a cordon signal of the PN. The action step duration is set at 5 seconds, drawing on methodologies from several existing adaptive traffic signal control studies (Chu et al., 2019; Sun and Yin, 2018; Yu et al., 2023). Agents are required to select actions based on current, real-time states every 5 seconds. This high frequency of interaction between agents and the environment ensures the capture of the most recent traffic state, and this communication is achievable in a connected vehicle environment (Sun and Yin, 2018). The minimum duration for a green light is set at 10 seconds. Additionally, the duration of the yellow light from previous actions is recorded and factored into the agent's decision-making to avoid the waste of green time due to frequent phase changes. Furthermore, agents are acting locally, while the global instructions can be updated at a different pace. More comprehensive details on this will be discussed later in the paper. Following the methodology outlined in Yu et al. (2023), the definitions of the agent's actions, states, and rewards are specified below.

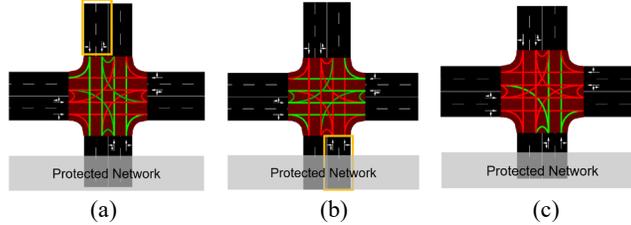

Fig. 3. Phases of cordon signals. (a) Phase 1, (b) phase 2, and (c) PC phase.

a) Agent's action

The action at step $t$ is a non-negative integer, $a_t = 0, 1, 2, \cdots$. It represents the index of the predefined signal phase that will be activated at the next step, and the last index is the PC phase, which is green for the outflow of PN only. Taking the signalized intersection shown in Fig. 3 as an example, $a_t = 0, 1, 2$ represents activating phase 1, phase 2, and PC phase in the next action step, respectively. If the two subsequent actions are different, the yellow and all-red phase of 5 s will be activated.

b) Agent's states

Intersection $i$ is the signalized intersection that agent $i$ controls. The agent's state consists of several elements:
- The summation of the total stopping time of all incoming legs of intersection $i$ during the last action step $t$, $D_{t,i}$;
- The set of occupancies of all incoming legs of intersection $i$ at step $t$, $[O_{t,i,l}], \forall l \in L_i$, where $L_i$ is the set of incoming legs of intersection $i$;
- The total number of yellow phases of agent $i$ in the last ten actions $y_{t,i}$; and
- The set of actions taken by agent $i$ and its direct neighbors at the last action step.

The occupancy of a link is defined as the total length of vehicles divided by the link length. The actions of direct neighbors are recorded in the state as the agents are supposed to learn to cooperate.

c) Agent's reward

The agent's reward consists of three sub-rewards aiming to reduce the traffic delay and prevent extreme scenarios (e.g., spillback, excessive phase switches) from happening. These sub-rewards are calculated separately and summed up to get the final reward at this action step. The agent's reward can be summarized as Eq. (4)-(7).

$$R_{t,i} = r'_{t,i} + \sum_{\forall l \in L_i} r''_{t,i,l} + r'''_{t,i} \tag{4}$$

where $R_{t,i}$ is the reward of agent $i$ at action step $t$. $r'_{t,i}$, $r''_{t,i,l}$, and $r'''_{t,i}$ are sub-rewards which are calculated as below:

$$r'_{t,i} = \begin{cases} 1 & D_{t,i} < D_{t-1,i} \\ -1 & D_{t,i} \geq D_{t-1,i} \end{cases} \tag{5}$$

$$r''_{t,i,l} = \begin{cases} 0 & O_{t,i,l} < \hat{O} \\ -1 & O_{t,i,l} \geq \hat{O} \end{cases} \tag{6}$$

$$r'''_{t,i} = \begin{cases} 0 & y_{t,i} \leq \hat{y} \\ -1 & y_{t,i} > \hat{y} \end{cases} \tag{7}$$

where $\hat{O}$ and $\hat{y}$ are the critical occupancy and the critical number of yellow phases in the last ten actions, respectively. $r'_{t,i}$ aims to shorten the total stopping time step by step. $r''_{t,i,l}$ is utilized to prevent the incoming legs from being oversaturated. $r'''_{t,i}$ monitors the waste of actual green time due to phase switches.

*3.3. Physics regularization for action-value function*

In the testing process of the MARL framework, the global feedback from the PC is integrated with local agents' action-value functions to regularize the agents' behaviors. Local control agents make decisions at every action step in the testing process by following a specific set of steps *a* to *d* shown in Fig. 4. The reward feedback in process *e* is necessary for the training process to update the weights of ANN, but it is not required in the testing process since no update in ANN is needed. In process *b*, the ANN evaluates and outputs the Q-values of all actions with the current state. Then, the action with the maximum Q-value is chosen in process *c*. Without affecting ANN's estimation, the PC feedback is integrated with these Q-values before the final action is selected in process *c*. Integrating an exogenous factor (i.e., PC feedback) between processes *b* and *c* will not violate the integrity of ANN functionality since the ANN solely works in process *b* and remains unaltered in the testing process. The MARL agents are trained to converge to the optimal or suboptimal control policy from the local perspective. This process enables regularizing the agent's decision with the global network state from the PC during testing.

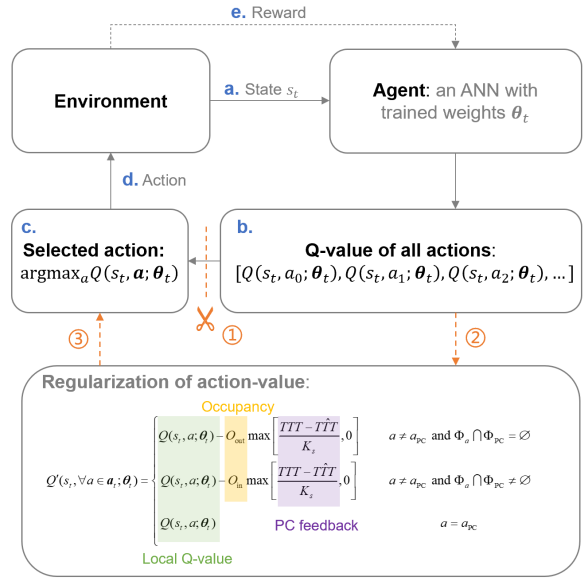

Fig. 4. Decision-making process in a value-based RL algorithm.

Assuming a simulated environment based on microscopic traffic, MFD is chosen to support the PC strategy and provide global feedback. No PC strategy is needed when the current TTT inside the PN is lower than the critical TTT. When the current TTT is higher than the critical TTT, all signals on the cordon of PN will receive the PC feedback $\left[ TTT - T\hat{T}T \right]$, which is the difference between the current observation of TTT and the critical TTT. When the action of the local agent with maximal value is not the PC phase, the local Q-values of these alternative actions are negatively affected by the PC feedback. To address the trade-off between PC feedback and action values of local agents, a weight parameter $K_s$ is introduced, which needs to be calibrated with the testing network. The occupancy of some specified legs of the cordon intersections also contributes to regularizing the final action-value function of the agents, which is explained in the following.

Regarding the update of global and local information, we designed an asynchronous method to integrate PC feedback with RL agents. In the asynchronous method, the PC feedback is updated every signal cycle. Given the absence of a predefined cycle length in our proposed adaptive signal control strategy, we have adopted a 70-second cycle length from our fixed control benchmarks. We chose to align the update pace of PC feedback with the cycle length to maintain consistency with most existing PC strategies (Keyvan-Ekbatani et al., 2015a; Keyvan-Ekbatani et al., 2015b; Ni and Cassidy, 2019; Su et al., 2023; Zhou and Gayah, 2023). Meanwhile, updates for the Q-values of the MARL agents and the occupancy rates of relevant legs occur every 5 seconds. Consequently, the global state of the network (i.e., TTT within the PN) that the MARL agents receive is not updated as frequently as local traffic states, leading to potential information delays. To evaluate the potential performance losses associated with these delays in the asynchronous method, a synchronous method is also conducted for comparison. The update paces of the PC feedback, Q-values, and leg occupancy are consistently set at 5 seconds in the synchronous method.

As Fig. 3(c) shows, the PC phase is configured with green for outflow movements of PN and red for all other movements. This phase may decrease local traffic efficiency at the intersection level, as some non-conflicted movements of outflow are not permitted, reducing the throughput of this intersection. However, the PC phase and other local control phases are not always completely opposing. An example is shown with Figs. 3(a) and 3(b). For the sake of simplicity, the straight flow volume is assumed to be much larger than the turning flows in this example. This assumption will be relaxed later in this section. Regarding phase 1 in Fig. 3(a), if the occupancy of the *inflow leg* (highlighted by the yellow box) is low, the effects of activating the PC phase and phase 1 are similar. Similarly, regarding phase 2 in Fig. 3(b), if the occupancy of the *outflow leg* (highlighted with the yellow box) is low, the effects of the PC phase and phase 2 are also comparable, while the intersection throughput of phase 2 is much higher than the PC phase. Therefore, when integrating the local action value with PC feedback, the occupancy of some specific legs should be accounted for. To summarize this rule, Eq. (8) defines the physics-regularized action-value function.

$$Q'(s_t, \forall a \in a_t; \theta_t) = \begin{cases} Q(s_t, a; \theta_t) - O_{out} \max\left[\dfrac{TTT - T\hat{T}T}{K_s}, 0\right] & a \neq a_{PC} \text{ and } \Phi_a \cap \Phi_{PC} = \varnothing \\ Q(s_t, a; \theta_t) - O_{in} \max\left[\dfrac{TTT - T\hat{T}T}{K_s}, 0\right] & a \neq a_{PC} \text{ and } \Phi_a \cap \Phi_{PC} \neq \varnothing \\ Q(s_t, a; \theta_t) & a = a_{PC} \end{cases} \quad (8)$$

where $Q'(s_t, a_t; \theta_t)$ is the regularized Q-value. $a_{PC}$ is the action that activates the PC phase. $\Phi_a$ represents the set of movements that get green in the phase activated by action $a$. $\Phi_{PC}$ is the set of green movements in the PC phase. With the assumption that the straight flow volume is much larger than the turning flow, the movements in $\Phi_a$ only contain straight movements. $O_{out}$ is the occupancy of the *outflow leg*. $O_{in}$ is the occupancy of the *inflow leg*. $K_s > 0$ is the weight parameter of PC feedback reward that needs to be calibrated according to the scale of PN. $\max\left[\dfrac{TTT - T\hat{T}T}{K_s}, 0\right]$ ensures that the PC will be activated only when the observed TTT is larger than the critical TTT.

To relax the assumption that the through flow is much larger than the turning flow, the modified action-value function needs to be redefined with the occupancy of each movement. $M_{in}$ and $M_{out}$ are the set of movements that allow vehicles to get into or out of the PN, as shown in Fig. 5.

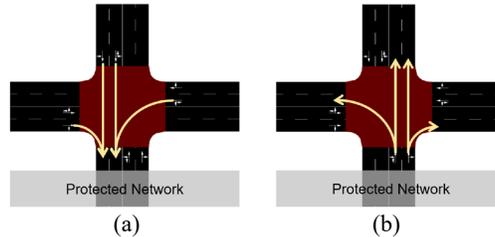

Fig. 5. Set of (a) inflow and (b) outflow movements of the PN.

When PC is needed, but the activated phase $a$ is not the PC phase, $M_{out,a}^{overlap} = M_{out} \cap \Phi_a$ is introduced as the overlap of outflow movements and green movements in phase $a$. $M_{out,a}^{com} = M_{out} - M_{out,a}^{overlap}$ is the complementary set of $M_{out,a}^{overlap}$ in $M_{out}$, which is the set of partial outflow movements that are not included in $\Phi_a$. Similarly, regarding the inflow movements, $M_{in,a}^{overlap} = M_{in} \cap \Phi_a$ represents the overlap of inflow movements and green movements in phase $a$. $M_{in,a}^{com} = M_{in} - M_{in,a}^{overlap}$ is the complementary set of $M_{in,a}^{overlap}$ in $M_{in}$. Then, the modified action-value function can be concluded in Eq. (9). The first and second lines in Eq. (8) are combined as the first line in Eq. (9).

$$Q'(s_t, \forall a \in \boldsymbol{a}_t; \boldsymbol{\theta}_t) = \begin{cases} Q(s_t, a; \boldsymbol{\theta}_t) - \left( \sum_{m_1 \in M_{out,a}^{com}} O_{m_1} + \sum_{m_2 \in M_{in,a}^{overlap}} O_{m_2} \right) \times \max\left[ \frac{TTT - T\hat{T}T}{K_s}, 0 \right] & a \neq a_{PC} \\ \\ Q(s_t, a; \boldsymbol{\theta}_t) & a = a_{PC} \end{cases} \quad (9)$$

where the $O_m$ is the occupancy of vehicles belonging to movement *m* on their origin link.

## 4. Simulation tests

*4.1. Test settings*

The proposed PC method is compared to different signal control strategies, including non-PC strategies and PC strategies, which are introduced below. These strategies only instruct signals located at the perimeter.

Non-PC strategies
- Fixed control: a two-phase fixed control. The cycle length is 70 seconds with equal green duration (30 seconds) for each phase and a 5-second yellow and all-red phase for each phase transition.
- Max pressure: the adaptive and decentralized traffic signal control proposed by (Varaiya, 2013). We choose this benchmark in case fixed control may not work well with heterogeneous and fluctuating traffic demands.

PC strategies:
- Bang-bang like feedback control (denoted as feedback control): the controller closes all gates for the following action step (5 seconds) when real-time TTT within the PN is higher than the critical TTT. It is a bang-bang like control strategy.
- Classical PI control: the PC strategies proposed by Keyvan-Ekbatani et al. (2012), in which the controller defines the total inflow of the PN and distributes the total flow proportionally to each gate according to its saturated flow rates.
- PI-cordon queue: the PC strategies proposed by Keyvan-Ekbatani et al. (2021). This strategy defines the total inflow of the PN based on the classical PI control and distributes the inflow of each gate according to the queue length outside the gate. The idea is to allow a larger metering rate for the gate with a larger relative queue length. The formulations of PI-cordon queue are detailed in Appendix A.1. Notely, a similar PC strategy is proposed to distribute the gate inflow considering the queue length outside the gate by Tsitsokas et al. (2023). Both the two strategies address the same objective as this study. We only chose the one proposed by Keyvan-Ekbatani et al. (2021) to conduct as a benchmark due to their similarity.

The proposed strategies:
- RL-ASYNC: the proposed method with asynchronous update paces (70 seconds for PC feedback and 5 seconds for MARL agents).
- MARL: the proposed method with the same update pace (5 seconds) of PC feedback and MARL agents.
- MARL-local: the proposed method without physics regularization from the PC strategy. Only local signal control is considered in this method. We conduct this method for the ablation study.

All the simulations are performed with the microscopic simulator SUMO (Alvarez Lopez et al., 2018). More details of the test settings are described below.

**1) Network and signal settings**

The proposed PC strategies and benchmarks are tested on a two-region network. To improve the simulation efficiency of the microscopic simulator SUMO, and focus on the queue length evolution outside the PN, the test network is simplified as shown in Fig. 6(a). The grey area, consisting of a 5×5 grid network, represents the PN, while the blue area is the region outside the cordon. The network in the outside region is simplified as 20 long links, connected to each cordon signal (i.e., the gate of the PN). The displayed numbers in Fig. 6(a) are the gate link IDs. The traffic interactions (e.g., traffic delay at intersections, link queue spillbacks, and gridlocks) within the outside region are disregarded as the traffic load of the outside region is not as large as the PN, and the oversaturated situation does not take into account for the outside region. The performance loss of the outside region due to the PC is evaluated by the queue length and traffic delay of the gate links, which are valuable indicators to show the completion of the goal of this study. All links in the network feature two lanes. The gate links outside the PN are 1000 meters long, while the link length inside the PN is 300 meters. All signals inside the PN are controlled by a two-phase fixed control in which the cycle length is 70 seconds, with equal green duration (30 seconds) for each phase and a 5-second yellow and all-red phase for phase transitions.

Fig. 6(b) displays the MFD for TTD and TTT inside the PN, where TTT and TTD are calculated over a time interval of 20 seconds. This means that the observed TTT and TTD represent the cumulative time spent and travel distance of all vehicles inside the PN during the last 20 seconds. The TTT range of 15000 to 20000 veh·s provides the maximum TTD, and thus 17000 veh·s is selected as the critical TTT. Previous studies have shown that the observed maximum TTT inside the PN for both classical PI and PI-cordon queue strategies slightly exceeds the critical TTT. This may be due to the delayed effectiveness of implementing an action after observing the network state. Similarly, in the proposed PC strategy, the observed maximum TTT is also expected to be larger than the critical TTT since the strategy doesn't force all gates to apply perimeter control. However, our simulation tests show that the feedback control in benchmarks keeps the maximum TTT inside the PN exactly around the critical TTT due to its direct, immediate, and forced action policy to all gates. This is in contrast to other PC strategies and may result in worse performance outside the cordon due to the stricter metering policy. To conduct fair comparisons among the benchmarks, the critical TTT of feedback control is modified to 20000, which is similar to the observed maximum TTT in other benchmarks. Appendix B displays the performance comparisons between feedback control with critical TTT of 17000 veh·s and 20000 veh·s, and suggests that the latter achieves better global performance.

There are three predefined phases for each cordon signal in the proposed PC strategy, as shown in Fig. 3. Phase 1 is green to the two legs for transfer flow (inflow and outflow) of PN. Phase 2 is green for the two legs parallel to the edge of the PN cordon. Phase 3 is the PC phase in which only the outflow leg gets green.

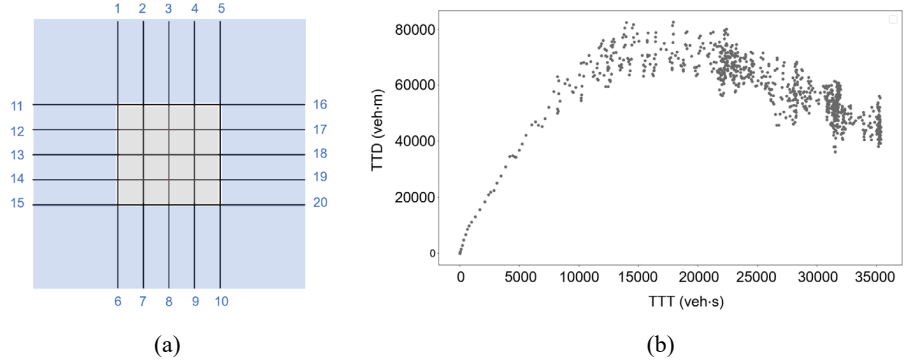

Fig. 6. (a) Testing network and (b) MFD of the PN.

**2) Demand settings**

Since the proposed method aims to address the unbalanced traffic demand by heterogeneous cordon signal behavior, two sets of demand are designed to get different unbalanced demand patterns, in which one is unbalanced from cordon edge to edge (demand 1) and the other one is unbalanced from gate to gate (demand 2). The demand settings are displayed in Table 1. For each simulation, the traffic is generated during the first 4800 seconds. Within the 4800 seconds, the demand values and traffic distributions change randomly every 1200 seconds to create the fluctuated traffic demand. All the simulations continue running after 4800 seconds until all vehicles arrive at their destinations. The intervals provided in the table represent demand ranges within which the exact demand values are randomly generated. To generate an unbalanced demand in different directions of the cordon, various levels of demand ranges are allocated to different origins. In demand 1, the demand levels in the four directions are different and fluctuated during the 4800s, also shown in Fig. 7. The shadow in Fig. 7 represents the random range. The demand range for the gates in one direction is the same, while the exact demand volumes vary due to the randomness. In demand 2, the notation "Shuffle[·]" represents shuffling the elements inside the brackets randomly. Five gate links in the same direction of the cordon are assigned with five shuffled demand ranges in one set, ensuring an unbalanced traffic distribution among gates and along the entire cordon. The demand flow distribution is also renewed every 1200s by re-shuffling the ranges. We can not display demand 2 in the same way as Fig. 7 since the shuffling process is different according to the random seed. In every training and test episode, demand is generated with different random seeds. In both demand patterns, the destinations of these demands are located both outside and inside the PN with random turning. 27% of the demand starts from the gate link and ends inside the PN, 63% starts from the gate link and ends outside of the PN, and 10% starts and ends inside the PN, with the origin being the downstream link of the gate. The traffic demand loads we set in this study have been tested with fixed control to ensure the throughput drop of MFD occurs, and meanwhile, the traffic is not too heavy to cause permanent gridlocks to the fixed control system.

Table 1. Settings of demand ranges (vehs/h)

| Origin (gate link ID) | Time interval (demand 1) | | | | Time interval (demand 2) | | | |
|---|---|---|---|---|---|---|---|---|
| | 0-1200s | 1200-2400s | 2400-3600s | 3600-4800s | 0-1200s | 1200-2400s | 2400-3600s | 3600-4800s |
| North (1 to 5) | 1200-1440 | 1080-1320 | 960-1140 | 810-990 | Shuffle[680-820, 810-990, 960-1140, 1080-1320, 1200-1440] | | | |
| South (6 to 10) | 810-990 | 960-1140 | 1080-1320 | 1200-1440 | Shuffle[810-990, 680-820, 1080-1320, 1200-1440, 960-1140] | | | |
| West (11 to 15) | 960-1140 | 1200-1440 | 810-990 | 1080-1320 | Shuffle[960-1140, 1080-1320, 1200-1440, 680-820, 810-990] | | | |
| East (16 to 20) | 1080-1320 | 810-990 | 1200-1440 | 960-1140 | Shuffle[1080-1320, 1200-1440, 680-820, 810-990, 960-1140] | | | |

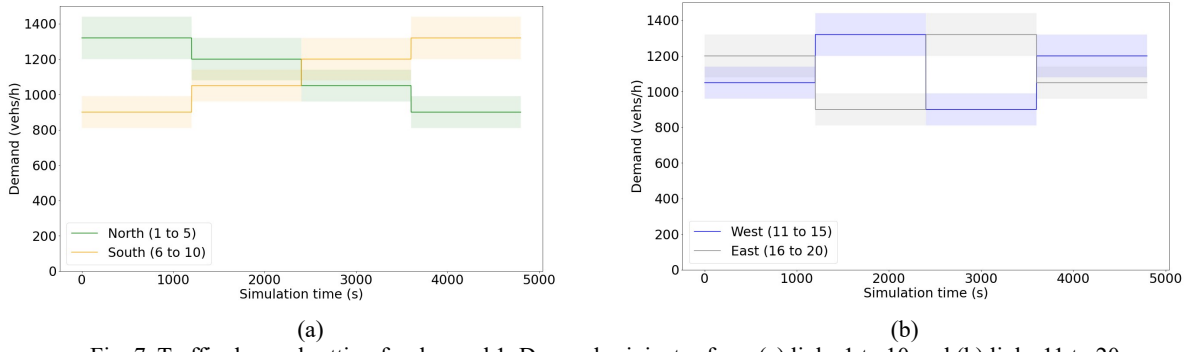

Fig. 7. Traffic demand setting for demand 1. Demand originates from (a) links 1 to 10 and (b) links 11 to 20.

**3) Relevant parameters**

The simulation involves vehicles of a 5-meter length with a minimum gap of 2.5 meters. The critical occupancy of each link is 75% of the full occupancy (Lo et al., 2001), which is 0.5 in this study. The speed limitation is 50 km/h. The critical number of phase switches in the last ten action steps is 2, the same as the number of phases in one cycle of fixed control (Yu et al., 2023).

The structure of ANN refers to (Vidali, 2021), where the number of layers and layer width are 4 and 400, respectively. The discount factor of future reward is 0.95, and the learning rate is 0.001. Several values ranging from 250 to 1500 with a step length of 250 have been tested for the weight parameter $K_s$, and we kept the one leading to the minimum TTT of the entire network, which is 750 in this testing network. Regarding classical PI control and PI-cordon queue, $K_P$ and $K_I$ are 2 and 0.5, respectively, according to grid search experiments.

**4) Training process**

The proposed MARL framework is distributed, and the agents are transferable. Only cordon signals on the North edge of PN are trained for 70 episodes. During the testing process, the trained agents are inherited by other untrained cordon signals. The exploration rate decreases linearly from 1 to 0.02 over the first 50 training episodes and remains unchanged later. At the end of the simulation in each episode, the ANN is trained for 800 epochs, and the replacement frequency of the target network is every 100 epochs. Demand 2 is utilized to generate demands during training, and both demand settings are tested with the same trained agents to demonstrate the proposed strategy's robustness. The reward curves of the five agents during training are shown in Fig. 8, and convergence is observed after 45 episodes. The episode in Fig. 8 starts from 8 as the window size of the moving average process is 8.

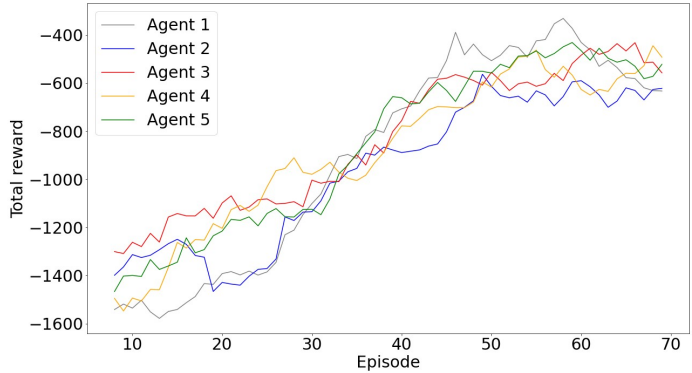

Fig. 8. Reward curves during the training process.

*4.2. Test results*

The TTT and cumulative TTD within the PN (denoted as PN-TTT and PN-TTD), the TTT of the Entire Network (denoted as EN-TTT), the total cordon queue length, and the carbon emissions are the Key Performance Indicators (KPIs) for the control performance evaluations. Both demand settings are tested with the same trained agents, and the KPIs for each control strategy with demand 1 and demand 2 can be found in Tables 2 and 3, respectively. The random seeds for demand generation in demand 1 and 2 are set to 15000 and 20000, respectively. There is no significant difference between the proposed MARL-ASYNC and MARL strategies. MARL-local fails to complete the simulation in the test of demand 1, as severe gridlocks that cause a standstill occur in the network during the simulation. Therefore, the KPIs of MARL-local in demand 1 are unbounded and meaningless, which are removed from Table 2. Detailed comparison analyses between the proposed strategies and benchmarks are followed.

Table 2. Control performance of each PC strategy in the test with demand 1 (random seed = 15000).

| Control Strategy | PN-TTT ($10^5$ veh·s) | PN-TTD ($10^6$ veh·m) | EN-TTT ($10^5$ veh·s) | Total cordon queue ($10^5$ veh·s) | Carbon emission ($10^6$ g) |
|---|---|---|---|---|---|
| MARL | **57.75** | **22.06** | **239.93** | **75.02** | 48.54 |
| MARL-ASYNC | 58.62 | 22.12 | 240.55 | 75.30 | **48.48** |
| MARL-local | - | - | - | - | - |
| Max pressure | 102.02 | 23.57 | 419.31 | 118.99 | 79.93 |
| Fixed | 102.10 | 22.67 | 373.86 | 108.26 | 73.73 |
| PI | 69.41 | 23.19 | 327.35 | 114.96 | 63.10 |
| PI-cordon queue | 79.50 | 24.63 | 457.73 | 203.34 | 97.48 |
| Feedback | 69.98 | 23.59 | 340.43 | 113.05 | 66.54 |

Note: the notation of '-' represents a meaningless value due to the standstill in the simulation.

Table 3. Control performance of each PC strategy in the test with demand 2 (random seed = 20000).

| Control Strategy | PN-TTT ($10^5$ veh·s) | PN-TTD ($10^6$ veh·m) | EN-TTT ($10^5$ veh·s) | Total cordon queue ($10^5$ veh·s) | Carbon emission ($10^6$ g) |
|---|---|---|---|---|---|
| MARL | 59.38 | 21.54 | 233.18 | 71.11 | 46.94 |
| MARL-ASYNC | **56.75** | **21.52** | **228.61** | **68.52** | **46.07** |
| MARL-local | 96.22 | 22.25 | 319.42 | 81.79 | 77.69 |
| Max pressure | 93.24 | 23.13 | 355.92 | 83.56 | 66.12 |
| Fixed | 100.36 | 23.06 | 370.38 | 113.10 | 73.34 |
| PI | 73.23 | 22.73 | 338.26 | 127.34 | 66.13 |
| PI-cordon queue | 80.67 | 24.14 | 495.66 | 226.67 | 106.47 |
| Feedback | 72.40 | 23.18 | 344.01 | 119.30 | 68.35 |

**1) TTT and cumulative TTD within the PN**

According to Tables 2 and 3, the proposed MARL strategy reduces the PN-TTT by 16.80% and 17.99% compared to the best performance from benchmarks (PI or feedback control) with demand 1 and 2, respectively. MARL-ASYNC exhibits equivalent performance as MARL. PI-cordon queue underperforms classical PI, as it leads to higher heterogeneity for traffic distribution inside the PN compared to classical PI, which is explained detailedly in Appendix A.3. Figs. 9 and 10 display the PN-TTT and cumulative PN-TTD over simulation under the two demand settings. With the same traffic demand, the proposed physics-regularized methods, MARL-ASYNC and MARL, take minimum time to complete the simulation in both demand settings. Performances of MARL-local fluctuate. It fails to protect the network in demand 1, while it takes a shorter time to complete the simulation than other benchmarks in demand 2. Fig. 9 demonstrates that all PC strategies can maintain the PN-TTT for around 20000 seconds. The curves of these control strategies end at different cumulative PN-TTD in Fig. 10 because each vehicle in turning flows chooses the fastest path based on the real-time network state when it enters the network (Alvarez Lopez et al., 2018), leading to different total route lengths. According to Fig. 10, the cumulative PN-TTDs of the proposed strategies, feedback control, and fixed control are larger than PI control and PI-cordon queue from 3000s to 7000s, indicating higher throughput. No significant difference is found between the cumulative PN-TTD of the proposed strategy and fixed control in most simulation time. Still, in the end, the proposed strategy performs better and ends with minimum total PN-TTD.

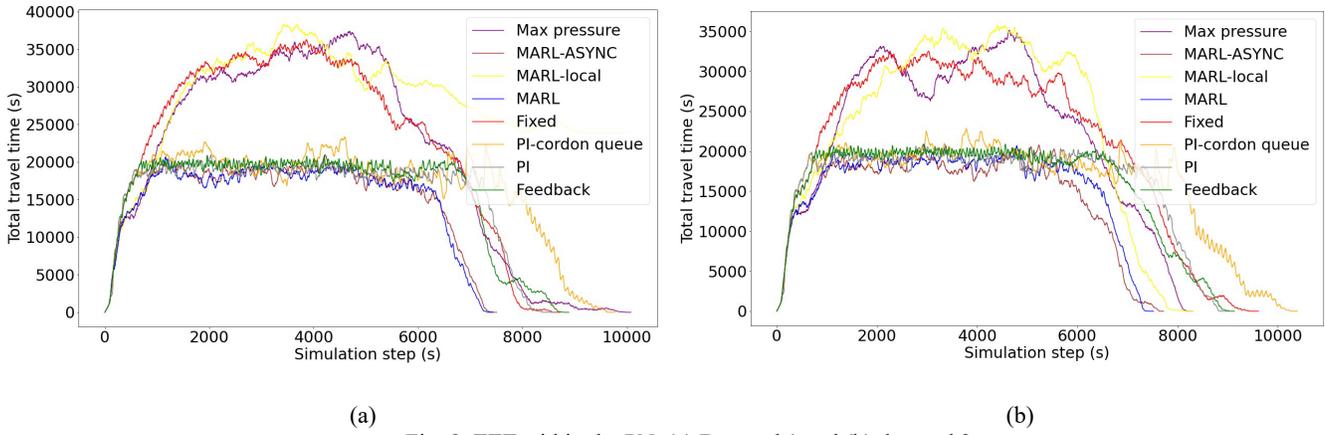

(a)            (b)

Fig. 9. TTT within the PN. (a) Demand 1 and (b) demand 2.

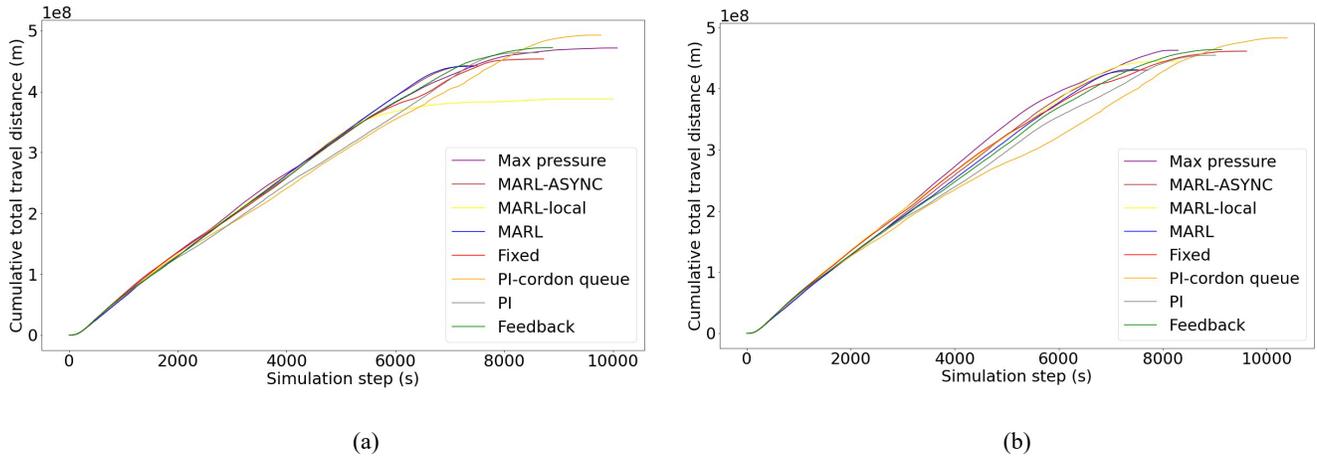

(a)            (b)

Fig. 10. Cumulative TTD within the PN. (a) Demand 1 and (b) demand 2.

**2) TTT, cumulative trip completion, and carbon emission of the entire network**

According to Tables 2 and 3, MARL outperforms benchmarks in the EN-TTT, showing a reduction of 26.71% and 31.06% compared to the best performance from benchmarks (classical PI control). MARL-ASYNC shows a similar performance on EN-TTT to MARL. In demand 2, the EN-TTT of MARL-local is less than all other benchmarks. Fig. 11 illustrates the EN-TTT for each control strategy with two demand settings. The feedback control and classical PI control show similar performance due to their comparable mechanisms. Fig. 11 demonstrates that the proposed physics-regularized strategies significantly enhance the network performance compared to benchmarks.

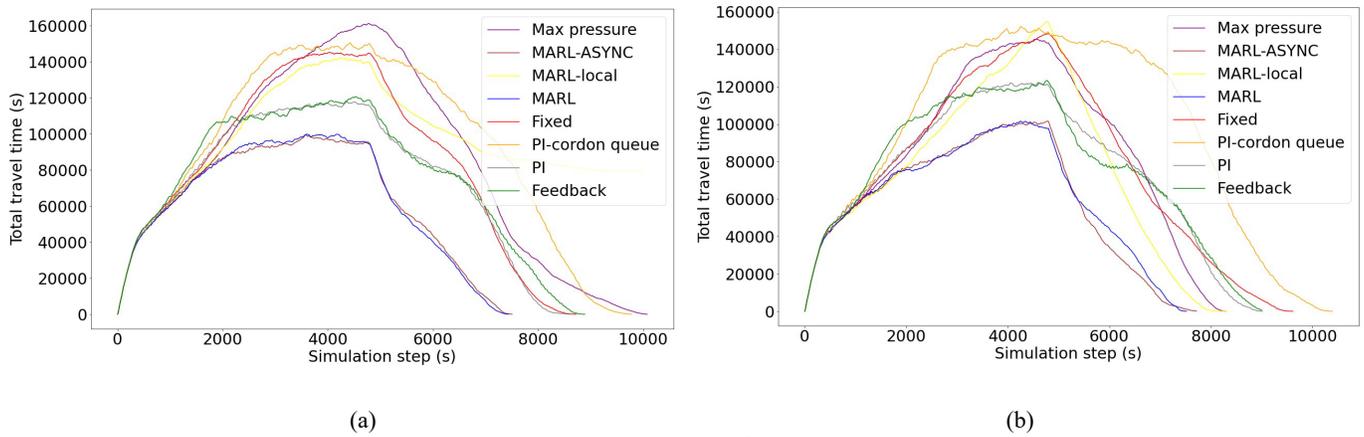

(a)            (b)

Fig. 11. TTT of the entire network. (a) Demand 1 and (b) demand 2.

Fig. 12 records the cumulative trip completion of the entire network. MARL-ASYNC and MARL show a higher efficiency of trip completion compared to other benchmarks in both two demand settings. The final cumulative completion of MARL-local in demand 1 is less than other methods as part of the vehicles are blocked in the network and not able to reach their destinations. Max pressure raises a similar trip completion rate with fixed control and PI control in the test of demand 2, but it underperforms these benchmarks in the test of demand 1.

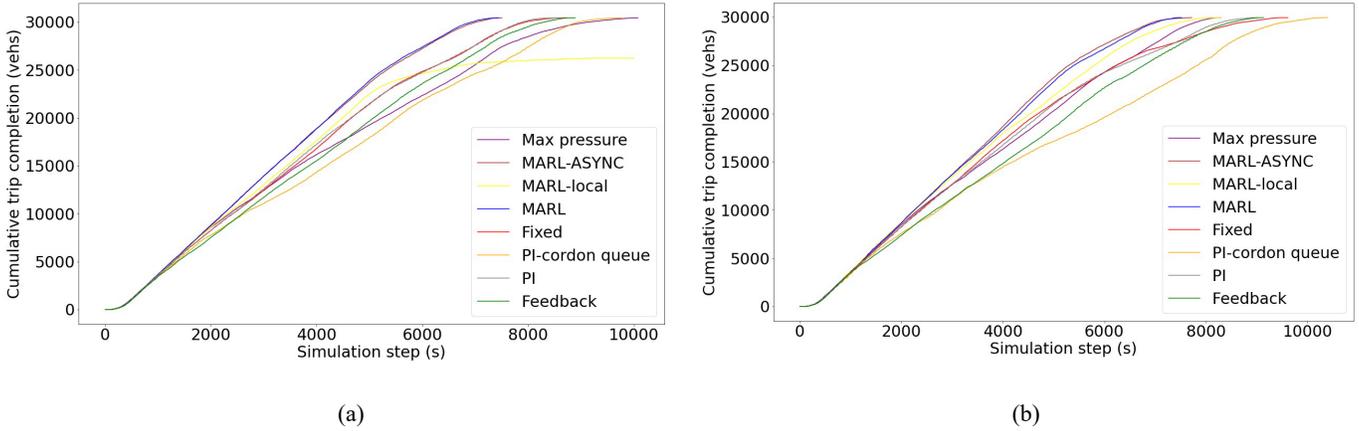

Fig. 12. Cumulative trip completion. (a) Demand 1 and (b) demand 2

Furthermore, Fig. 13 showcases the cumulative carbon emission throughout the simulation. The PI control, feedback control, and proposed strategy all reduced the cumulative carbon emission compared to fixed control. The carbon emission of PI-cordon queue is larger than fixed control due to the heterogeneous traffic distribution in PN that caused severe gridlock. MARL decreases the total carbon emission by 23.07% and 29.02% with demand 1 and 2, respectively, compared to the best performance in benchmarks. The curves of MARL-ASYNC are almost overlapped with MARL, indicating a similar reduction in carbon emissions. In demand 1, the cumulative carbon emission of MARL-local is unbounded.

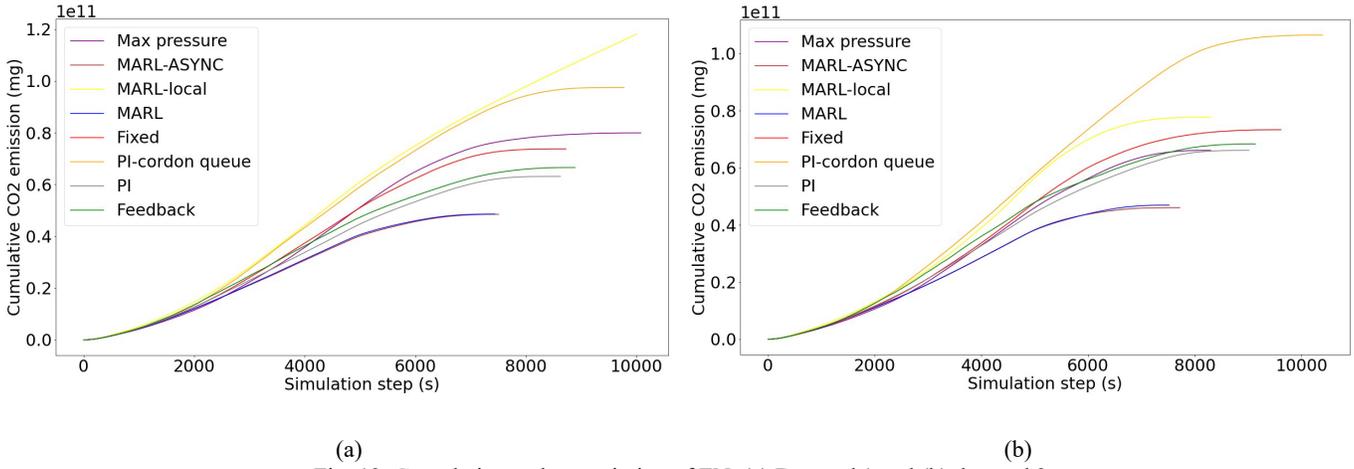

Fig. 13. Cumulative carbon emission of EN. (a) Demand 1 and (b) demand 2.

### 4) MFDs and cordon queue

Fig. 14 depicts the MFDs of MARL-based strategies, fixed control, and feedback control under demand 1, while the MFDs of PI control and PI-cordon queue are displayed in Appendix A.3. The black dots represent data collected up to 4800s, while the gray dots indicate data collected after 4800s. Traffic demand is generated only before 4800s in the simulation and remains zero thereafter. The gridlocks in the MFD of MARL-local are pointed out by the circle. The TTD falls to zero while the TTT remains high. The fixed control and max pressure strategies experience a significant drop in throughput, while the proposed physics-regularized strategies successfully maintain the TTD within the PN at its maximum level.

Additionally, Fig. 15 despicts the total cordon queue length evolution of each strategy during the simulation. The fixed control, max pressure, and MARL-local strategies initially have lower queue lengths than all other strategies due to the lack of a metering policy or physics regularization. However, as the accumulations within the PN increase in these three strategies, the cordon queue length becomes larger than PI control, feedback control, and the proposed physics-regularized strategies in the second half of the simulation. MARL reduces the total cordon queue length by 30.70% and 37.13% with demand 1 and 2, respectively, compared to the best performance from benchmarks. The large queue length in PI-cordon queue is due to the heterogeneous traffic distribution inside the PN, which has been explained in Appendix A.3.

Furthermore, it is crucial to understand the distributed traffic state of each gate link outside the cordon. Fig. 16 shows the average delay experienced by vehicles on each gate link before entering the PN, indicating that the proposed physics-regularized strategies effectively mitigate large delays (dark red blocks) and maintain a larger number of minor delays (blue blocks) for the gate links compared to benchmarks. The distributed cordon queue length evolution on gate links in the proposed strategies, feedback control, fixed control, and max pressure under demand 1 is depicted in Fig. 17. By incorporating the distributed cordon queue of PI control and the PI-cordon queue depicted in Appendix A.3, it is evident that the proposed physics-regularized strategies significantly reduce the queue length on the majority of gate links compared to benchmarks, allowing all queues to dissipate with a shorter time.

Overall, the proposed strategies considerably enhance both the global network performance and the distributed traffic delay of each gate link compared to benchmarks. There is no significant difference between the proposed MARL and MARL-ASYNC methods. The underperformance of MARL-local indicates the necessity of the physics regularization of the proposed MARL-based methods.

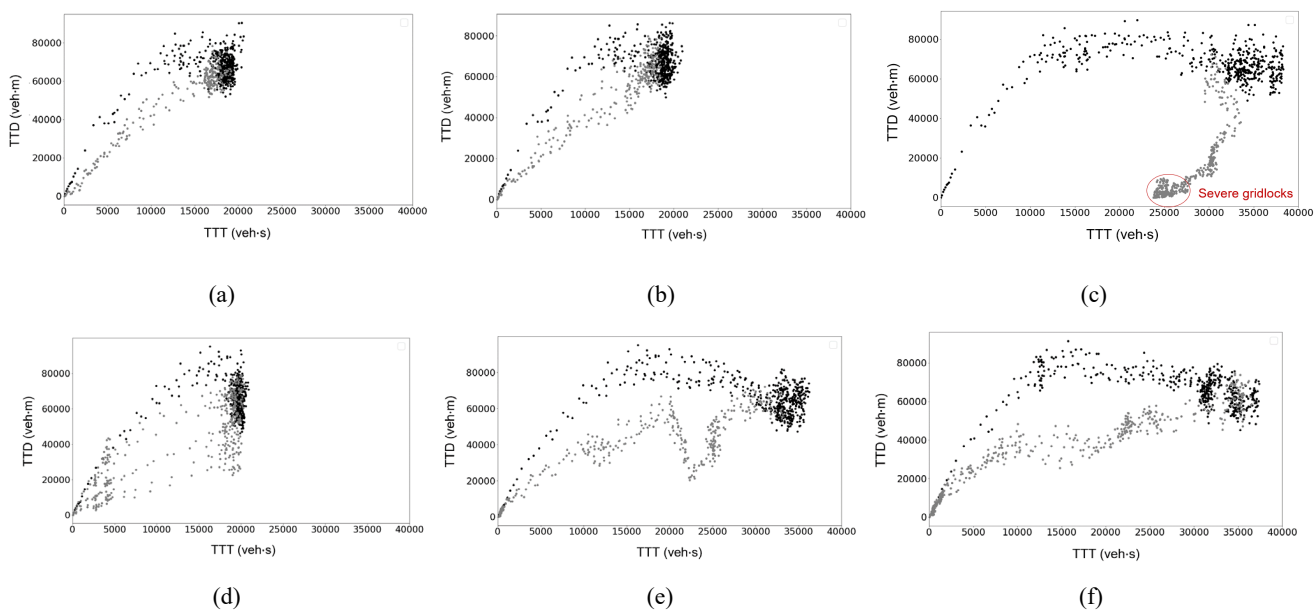

Fig. 14. MFD of each control strategy with demand 1. (a) MARL, (b) MARL-ASYNC, (c) MARL-local, (d) feedback control, (e) fixed control, and (f) max pressure.

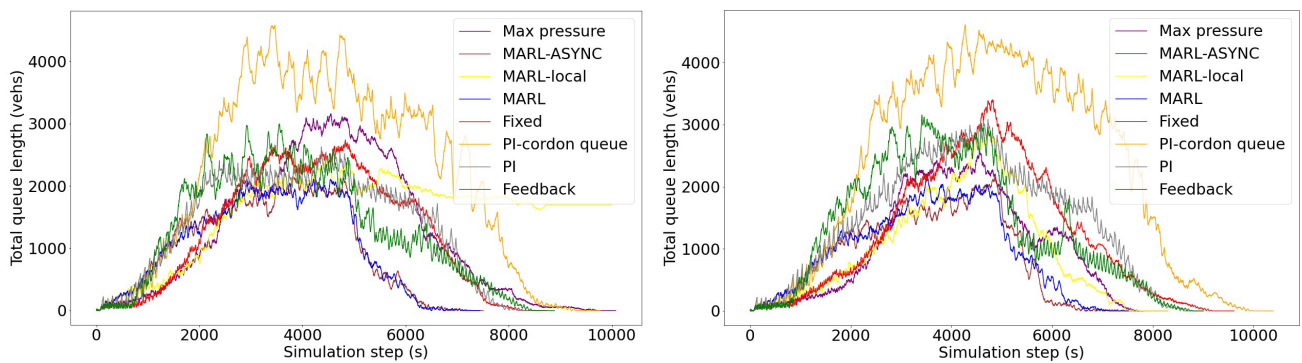

(a) (b)

Fig. 15. Total cordon queue length. (a) Demand 1 and (b) demand 2.

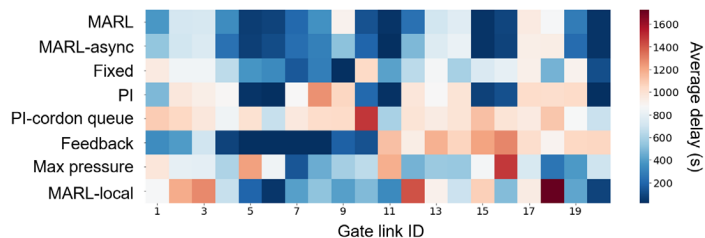

Fig. 16. Average delay of each gate link with demand 1.

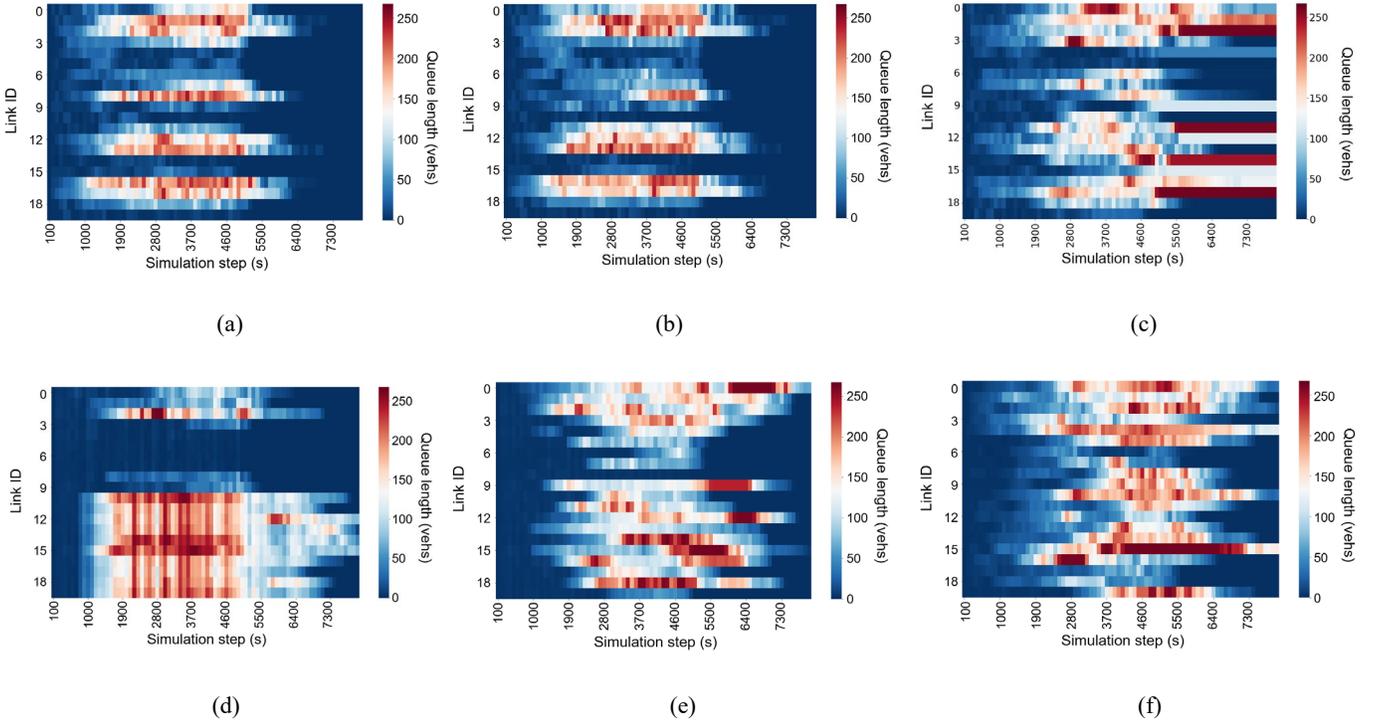

Fig. 17. Queue length evolution on each gate link with demand 1. (a) MARL, (b) MARL-ASYNC, (c) MARL-local, (d) feedback control, (e) fixed control, and (f) max pressure.

## 5. Conclusions

This paper introduces a Perimeter Control (PC) strategy utilizing a physics-regularized Multi-Agent Reinforcement Learning (MARL) framework. This strategy accounts for varying local traffic conditions along the perimeter by enabling heterogeneous metering rates for cordon signals. The approach is structured as a two-stage system: initially, the feedback PC issues global instructions at the network level, followed by the distribution of local control strategies for each cordon signal. The local control framework employs MARL agents to manage individual cordon signals independently. The physics-regularized MARL framework encodes knowledge from the Macroscopic Fundamental Diagram (MFD) into the action-value functions of the MARL agents, enabling local agents to be cognizant of the overall traffic state within the protected network. This distributed framework enhances the robustness and transferability of the strategy.

Simulation tests with different heterogeneous traffic demand patterns have demonstrated that the proposed strategy outperforms state-of-the-art feedback control strategies in increasing network throughput, reducing carbon emissions, and decreasing cordon queue length. The distributed performance of each gate link is also improved with the proposed method. The proposed strategy will be further explored in future work on multi-region and multi-modal networks. Additionally, we will explore the application of model predictive control within the physics-regularized MARL framework to enhance control effectiveness.

## Declaration of Competing Interest

The authors declare that they have no known competing financial interests or personal relationships that could have appeared to influence the work reported in this paper.

## Acknowledgments

This work was supported by Research Grants Council of the Hong Kong Special Administrative Region, China (Project No. PolyU/25209221) and the European Union's Horizon 2020 research and innovation program under Grant Agreement no. 953783 (DIT4TraM).

## Appendix A. Performance analysis of existing PI control strategies

*A.1 Formulation of PI-cordon queue*

Regarding the inflow distribution of PI-cordon queue in (Keyvan-Ekbatani et al., 2021), the objective of the integer quadratic programming problem is to minimize the sum of the squared relative queue length on gate links outside the cordon. The programming problem is formulated as Eq. (A.1).

$$\min \sum_{i=1}^{n} \frac{N_{rel,i}^2(k+1) N_{max,i}}{T} \quad \text{s.t.:} \sum_{i=1}^{n} q_i(k) = q_g(k) \text{ and } q_{min,i} \leq q_i(k) \leq q_{max,i} \quad (A.1)$$

where $T$ is the cycle length. $N_{rel,i}(k+1)$ is the relative queue length on gate link $i$ at time step $k+1$. $N_{max,i}$ is the maximum queue length on gate link $i$. $q_i(k)$ is the inflow of PN from gate $i$ during time step $k$ to step $k+1$. $q_{min,i}$ and $q_{max,i}$ are the lower and upper bounds of the leaving flow of gate link $i$.

*A.2 Theoretical analysis of the traffic distributions with classical PI and PI-cordon queue*

Store-and-forward paradigm is utilized to analyze the traffic distribution inside the PN (Aboudolas et al., 2009). The link dynamic is given by Eq. (A.2).

$$x_z(k+1) = x_z(k) + T[i_z(k) - s_z(k) + d_z(k) - u_z(k)] \quad (A.2)$$

where $x_z(k)$ is the accumulation of link $z$ at time step $k$. $i_z(k)$ and $u_z(k)$ are the inflow and outflow of link $z$ during time $kT$ to $(k+1)T$. $d_z(k)$ and $s_z(k)$ are the external demand and the exit flow within the link, respectively, which are assumed to be known (Aboudolas et al., 2009). The inflow and outflow are given by Eq. (A.3) and (A.4).

$$i_z(k) = \sum_{w \in I_z} t_{w,z} u_w(k) \quad (A.3)$$

$$u_z(k) = g_z(k) S_z \quad (A.4)$$

where $I_z$ the set of inflow links of link $z$. $t_{w,z}$ is the turning rate towards link $z$ from the link $w$. $g_z(k)$ is the green time ratio of link $z$. $S_z$ is the saturated outflow rate of link $z$. The calculation of $u_z(k)$ in Eq. (A.4) is a simplified formulation. The outflow rate is set to be the saturated flow rate during all the green period. This simplification may cause inconsistency between the theoretical analysis and results in microscopic simulation. Here, we provide the theoretical analysis first, and the microscopic simulations are conducted following.

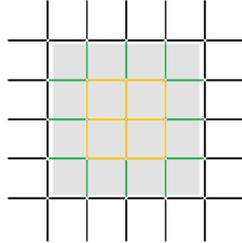

Fig. A1. Protected network in PC

Regarding the links that are not directly connected with the gates (depicted as yellow links in Fig. A1), no external demand or exit flow is set within these links in this study. Therefore, the accumulation evolution of these links is calculated as Eq. (A.5):

$$x_z(k+1) = x_z(k) + T\left[\sum_{w \in I_z} t_{w,z} g_w(k) S_w - g_z(k) S_z\right] \quad (A.5)$$

During the test, the settings of traffic demand (i.e. OD patterns) are kept the same under different control strategies. We assume that the turning rates of links inside the PN are consistent in the classical PI and PI-cordon queue. Thus, the distribution of traffic within the yellow links is the same in both strategies.

Regarding the links connected to the gates directly (depicted as green links in Fig. A1), the external demand and the exit flow are set to be proportional to the OD volumes which are heterogeneous random values. In this testing scenario, the traffic flows start or end within the PN are set to travel as long as possible to ensure the traffic activities inside the PN are effective. If the trip of vehicles ends as soon as they get into the PN, the traffic will dissipate soon and no congestion or oversaturated situation will occur. Therefore, only the inbound green links are equipped with external demand, while the external exit flow is set in the outbound green links. "Inbound" represents the direction from the cordon to the center of the PN, and "outbound" is the other direction. The accumulation evolutions of inbound and outbound links are shown as Eq. (A.6) and (A.7). The inflow of the gate connected with link $z$ is donated as $q_z(k)$.

$$x_z^{inbound}(k+1) = x_z^{inbound}(k) + T[i_z(k) - u_z(k) + d_z(k)] = x_z^{inbound}(k) + T[q_z(k) - g_z(k) S_z + p_d q_z(k)]$$
$$= x_z^{inbound}(k) + T[(1+p_d) q_z(k) - g_z(k) S_z] \quad (A.6)$$

$$x_z^{outbound}(k+1) = x_z^{outbound}(k) + T[i_z(k) - u_z(k) - s_z(k)] = x_z^{outbound}(k) + T[i_z(k) - u_z(k) - p_s i_z(k)]$$
$$= x_z^{outbound}(k) + T\left[(1-p_s) \sum_{w \in I_z} t_{w,z} g_w(k) S_w - g_z(k) S_z\right] \quad (A.7)$$

where the $p_d$ and $p_s$ are the proportions of external demand and the exit flow which are consistent for all relevant links.

According to Eq. (A.6), the accumulation of inbound link $z$ is related to inflow volume from the gate $q_z(k)$. In classical PI

control, the inflow volume from each gate is the same since their saturated flow rates are uniform. Thus, the distribution of traffic among the inbound links is homogenous. In PI-cordon queue, the inflow volume of each gate is different according to the queue length outside the gate, and the accumulations of inbound links are heterogeneous. Regarding outbound links, since the green time ratios are the same for these links, the distribution of traffic is also homogenous under both control strategies according to Eq. (A.7).

In conclusion, based on the store-and-forward paradigm, the distribution of traffic among outbound green links and inner yellow links is consistent under the classical PI and PI-cordon queue. The accumulations of inbound links are homogenous in the classic PI strategy while heterogeneous in PI-cordon queue. Therefore, the heterogeneity of traffic distribution inside the PN in PI-cordon queue is higher than that in the classical PI strategy, which may cause local gridlocks and violate the stability of the PN, resulting in PI-cordon queue being unfunctional in networks with dense inflow gates.

*A.3 Simulation comparisons between classical PI and PI-cordon queue*

Fig. A2(b) and 1(c) show the queue length evolution on each gate link outside the PN in the two control strategies. PI-cordon queue fails to protect the perimeter and causes severe queue spread compared to classical PI. The traffic state inside the PN was explored to understand this phenomenon. Figs. A3(a) and A3(b) show the number of vehicles on each link inside the PN under both strategies along the simulation and the traffic distribution at several simulation steps. The simulation results show a larger number of highly occupied links (dark red blocks) in PI-cordon queue than in classical PI control. This is due to PI-cordon queue allocating a large inflow to the gate with a large queue length, resulting in severe local congestion and even spillbacks to the connected links inside the PN. Meanwhile, the occupancy of links close to gates with low demand is relatively low due to the small volume of distributed inflow. From the geo-based heatmap in Fig. A3, severe congestion and unbalanced traffic distribution are found to occur in PI-cordon queue, which violates the stability of the network. Fig. A4(a) displays the standard deviation of vehicle numbers on each link inside the PN for the two control strategies. The standard deviation in PI-cordon queue is larger than that in classical PI during most simulation time, indicating that traffic distribution within PN is more unbalanced than in classical PI control. During the simulation steps from 2000s to 4800s (the duration of the demand generation except for warming-up time), the average standard deviations of classical PI control and PI-cordon queue are 11.44 and 15.17. PI-cordon queue increases the standard deviation by 32.60% compared to classical PI, leading to higher heterogeneity of traffic distribution inside the PN. The heterogeneity of traffic distribution causes the network throughput to drop even if the accumulation is still within the critical range (Ramezani et al., 2015). From the MFDs of classical PI control and PI-cordon queue (as depicted in Figs. A4(b) and A4(c)), the lower bound of TTD in the PI-cordon queue is lower than that in classical PI control around critical TTT, indicating that the stability of the network is violated in PI-cordon queue. Therefore, the classical PI control outperforms PI-cordon queue and stores fewer vehicles at the end of the simulation, as shown in Fig. A3.

This test demonstrates that the gate inflow distribution method in PI-cordon queue is unsuitable for dealing with the dense grid network in such a highly oversaturated scenario. It is crucial to refine the cordon gates' action with local implementation of the global PC strategy.

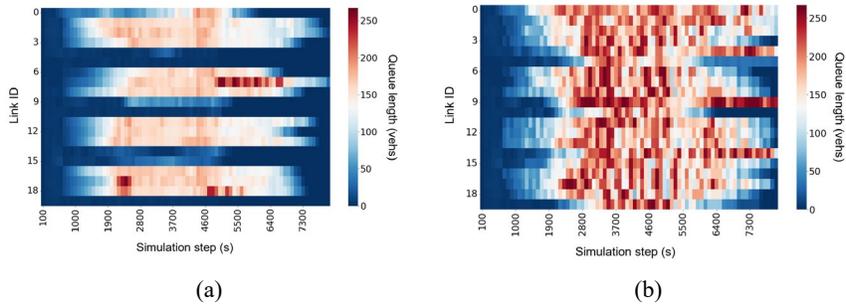

(a)                          (b)

Fig. A2. Control network and queue length evolution on each gate link. (a) classical PI control and (b) PI-cordon queue.

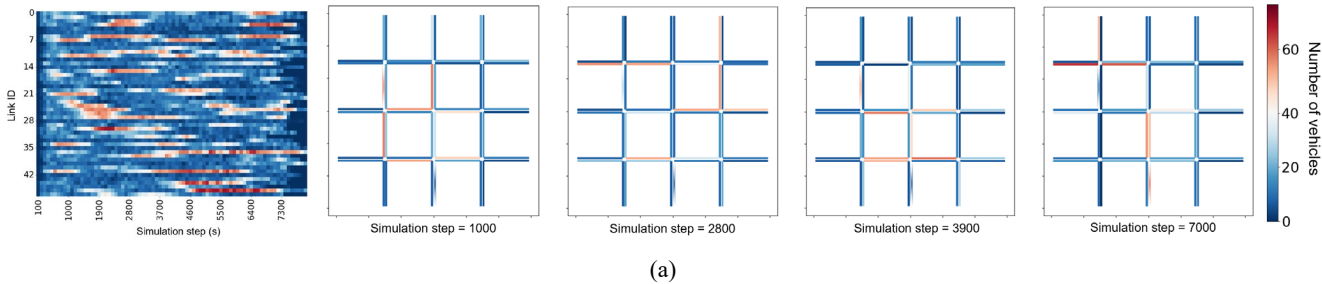

(a)

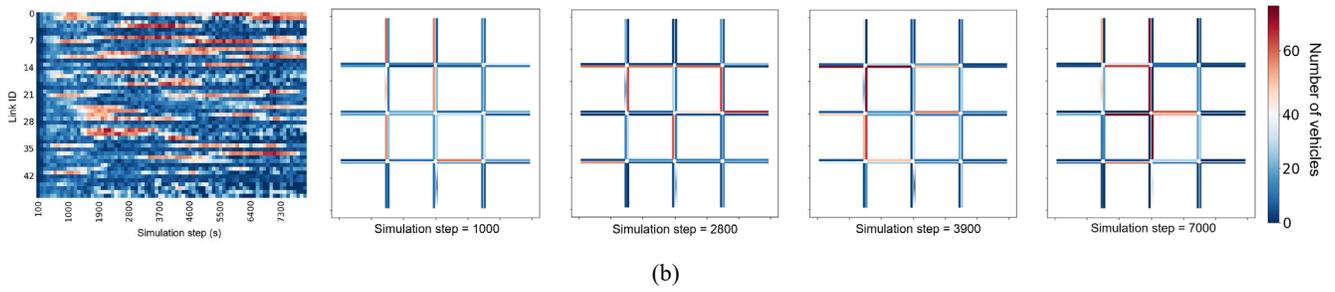

Fig. A3. Vehicle number evolution on each link inside PN. (a) Classical PI control, (b) PI-cordon queue.

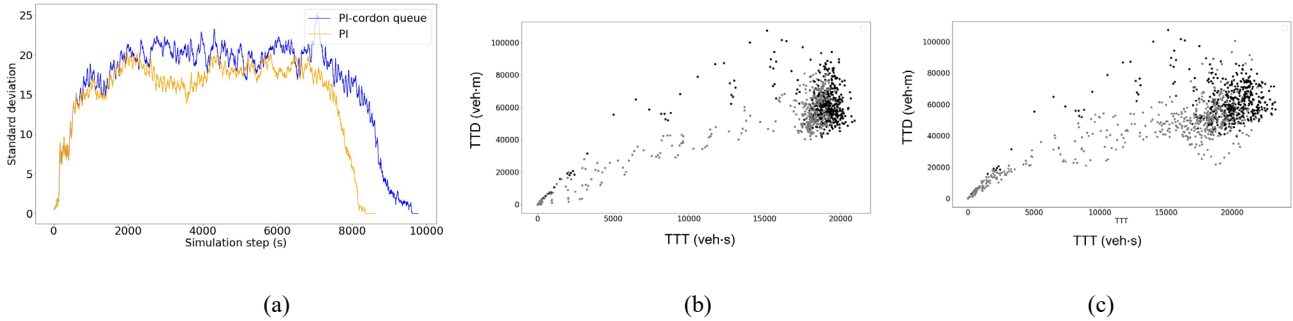

Fig. A4. Comparisons of network performance. (a) The standard deviation of vehicle numbers on each link inside PN, MFDs of (b) classical PI control, and (c) PI-cordon queue

## Appendix B: Performance comparison between feedback control with critical TTT = 17000 and 20000

According to the simulation tests with the setting demand 1, the feedback control with different critical TTT maintains the TTT inside the PN around the set critical TTT closely. Table B1 displays several overview indicators of the two tests. PN-TTT of the two tests is similar (as depicted in Fig. B1(a)) although the one with critical TTT of 17000 veh·s (denoted as "feedback-17000" in this section) maintains the PN-TTT in a lower level. Feedback control with critical TTT of 20000 veh·s (denoted as "feedback-20000" in this section) shows higher efficiency in outflow (as shown in Fig. B1(b)) which is represented by PN-TTD in the microscopic environment. Regarding the entire network, feedback-20000 achieves 8.27% smaller EN-TTT (as depicted in Fig. B2(a)) and a higher trip completion rate (as shown in Fig. B2(b)) compared to feedback-17000. The total queue length outside the cordon (Fig. B3(a)) and carbon emission of the entire network (Fig. B3(b)) in feedback-20000 are 19.70% and 10.97% lower than those in feedback-17000, respectively. In conclusion, feedback-20000 outperforms feedback-17000 in all indicators, and 20000 is selected to be the critical TTT of feedback control for the simulation tests in this paper.

Table B1. Control performance of feedback control with different critical TTT. (Demand 1, random seed = 15000).

| Critical TTT | PN-TTT ($10^5$ veh·s) | PN-TTD ($10^6$ veh·m) | EN-TTT ($10^5$ veh·s) | Total cordon queue ($10^5$ veh·s) | Carbon emission ($10^6$ g) |
|---|---|---|---|---|---|
| 17000 | 69.18 | 23.75 | 371.11 | 140.78 | 74.74 |
| 20000 | 69.98 | 23.59 | 340.43 | 113.05 | 66.54 |

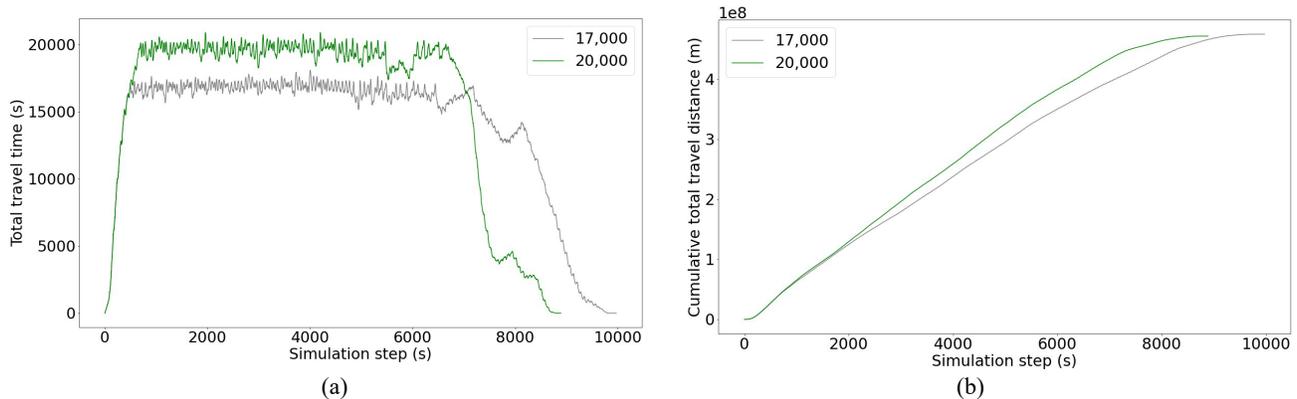

Fig. B1. (a) total travel time and (b) cumulative total travel distance of the PN.

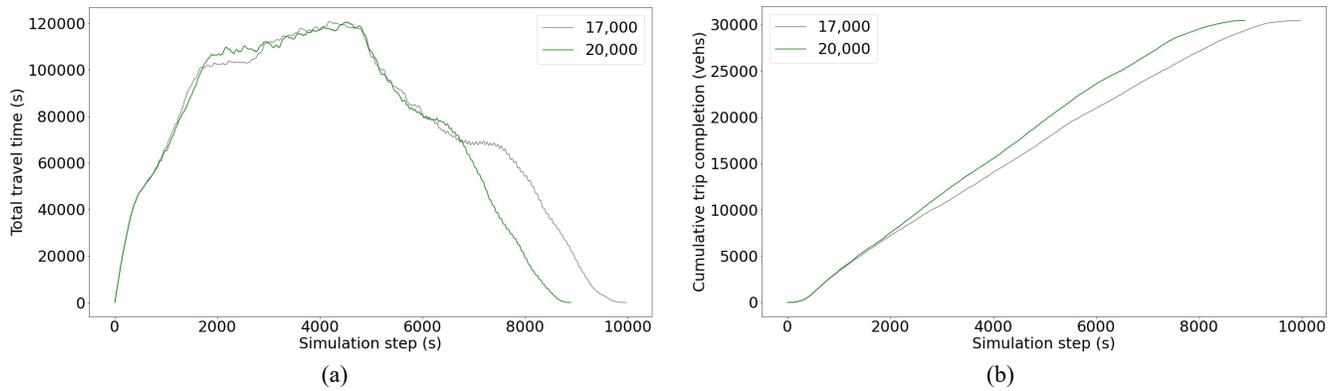

Fig. B2. (a) total travel time and (b) cumulative trip completion of the EN.

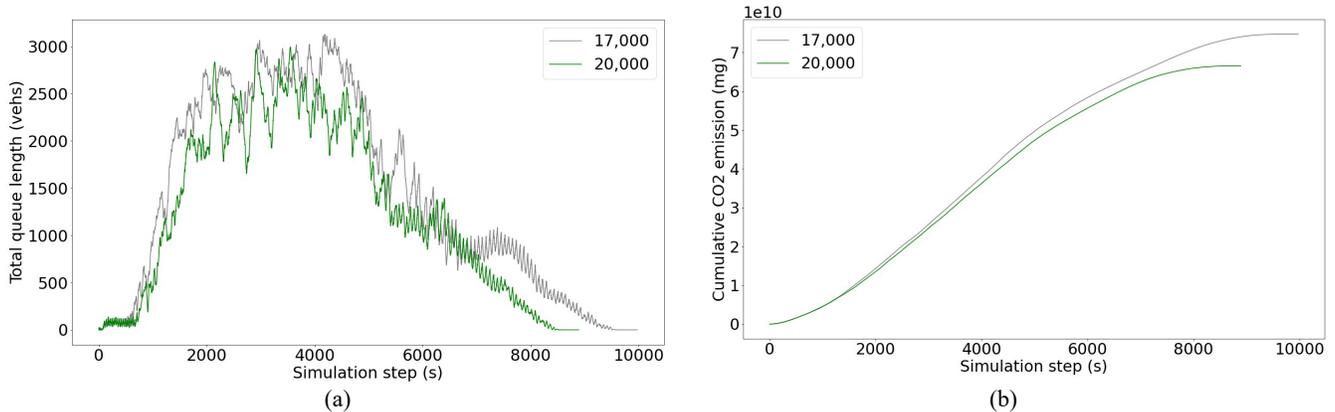

Fig. B3. (a) total queue length outside the cordon and (b) cumulative carbon emission of the EN.